\documentclass[letterpaper, 10 pt, conference]{ieeeconf}  
\usepackage{tikz}
\usepackage{amsmath}
\usetikzlibrary{matrix}
\usepackage{hyperref}

\IEEEoverridecommandlockouts                              

\title{\LARGE \bf
A Prototype System for High Frame Rate Ultrasound Imaging based Prosthetic Arm Control}

\author{Ayush Singh$^{1}$, Pisharody Harikrishnan Gopalkrishnan$^{1}$ and Mahesh Raveendranatha Panicker$^1$
\thanks{$^{1}$Ayush Singh, Pisharody Harikrishnan Gopalkrishnan and Mahesh Raveendranatha Panicker are with the Department of Electrical Engineering and Center for Computational Imaging, Indian Institute of Technology Palakkad, Kerala 678557, India {\tt\small correspondance: mahesh@iitpkd.ac.in}}%
}

\begin{document}

\maketitle
\thispagestyle{empty}
\pagestyle{empty}

\begin{abstract}
 The creation of unique control methods for a hand prosthesis is still a problem that has to be addressed. The best choice of a human-machine interface (HMI) that should be used to enable natural control is still a challenge. Surface electromyography (sEMG), the most popular option, has a variety of difficult-to-fix issues (electrode displacement, sweat, fatigue). The ultrasound imaging-based methodology offers a means of recognising complex muscle activity and configuration with a greater SNR and less hardware requirements as compared to sEMG. In this study, a prototype system for high frame rate ultrasound imaging for prosthetic arm control is proposed. Using the proposed framework, a virtual robotic hand simulation is developed that can mimic a human hand as illustrated in the link \cite{viddemo}. The proposed classification model simulating four hand gestures has a classification accuracy of more than 90\%. 
\newline

\indent \textit{Clinical relevance}—The proposed system enables an ultrasound imaging based human machine interface that can be a development platform for novel control strategies of a hand prosthesis.
\end{abstract}

\section{INTRODUCTION}
The inability to work, interact, and do daily duties independently after losing an upper limb can significantly reduce a person's quality of life. Artificial limbs are used to help patients who have lost limbs improve their function and quality of life. Despite substantial advances in prosthetic technology, rejection rates for sophisticated prosthetic devices remain high \cite{biddiss2007upper}. The majority of upper limb amputees discard their prosthetics due to their complexities and lack of intuitive control. Surface electromyography (sEMG), the most generally used method for controlling prosthetic arms, has severe limitations, including a low signal-to-noise ratio, poor amplitude resolution, and inability to capture complex muscle activations \cite{clancy2002sampling}. Also, the sensitivity to finer movement of muscles is very poor in the case of sEMG \cite{ortenzi2015ultrasound}. However, it is shown in literature that ultrasound imaging produces a much more detailed and dynamic mapping of muscles which can help in finer controls \cite{dhawan2019proprioceptive,engdahl2022first}. Also, the possibility to associate muscle patterns with specific finger movements using machine learning on ultrasound images has been presented in literature. In \cite{dhawan2019proprioceptive}, an ultrasound imaging based control with $5$ prosthetic users and $5$ able-bodied participants in a virtual target achievement and holding task for $5$ different hand motions was presented. In a recent work \cite{engdahl2022first}, the first demonstration of real-time prosthetic hand control using sonomyography to perform functional tasks was illustrated. \newline
However, a prototype system for end-to-end control of prostetic arm in a research environment hasn't been investigated before. In this work, a research ultrasound platform was employed to acquire the ultrasound images at high frame rates by employing multi-angle planewave ultrasound transmission \cite{bercoff2011ultrafast}. The proposed system was tested on healthy volunteers by having them do predefined hand motions while ultrasound images of their forearm muscle movements were captured using the Verasonics Vantage 128 Channel Research Ultrasound System. We constructed a training directory using the obtained ultrasound data and then used a supervised learning method based on k-nearest neighbours to predict individual finger movements. Participants performed hand motions in real-time while viewing a virtual hand simulation in the robotic operating system (ROS) Environment.
\begin{figure*}[ht]
    \centering
    \includegraphics[width=0.8\linewidth]{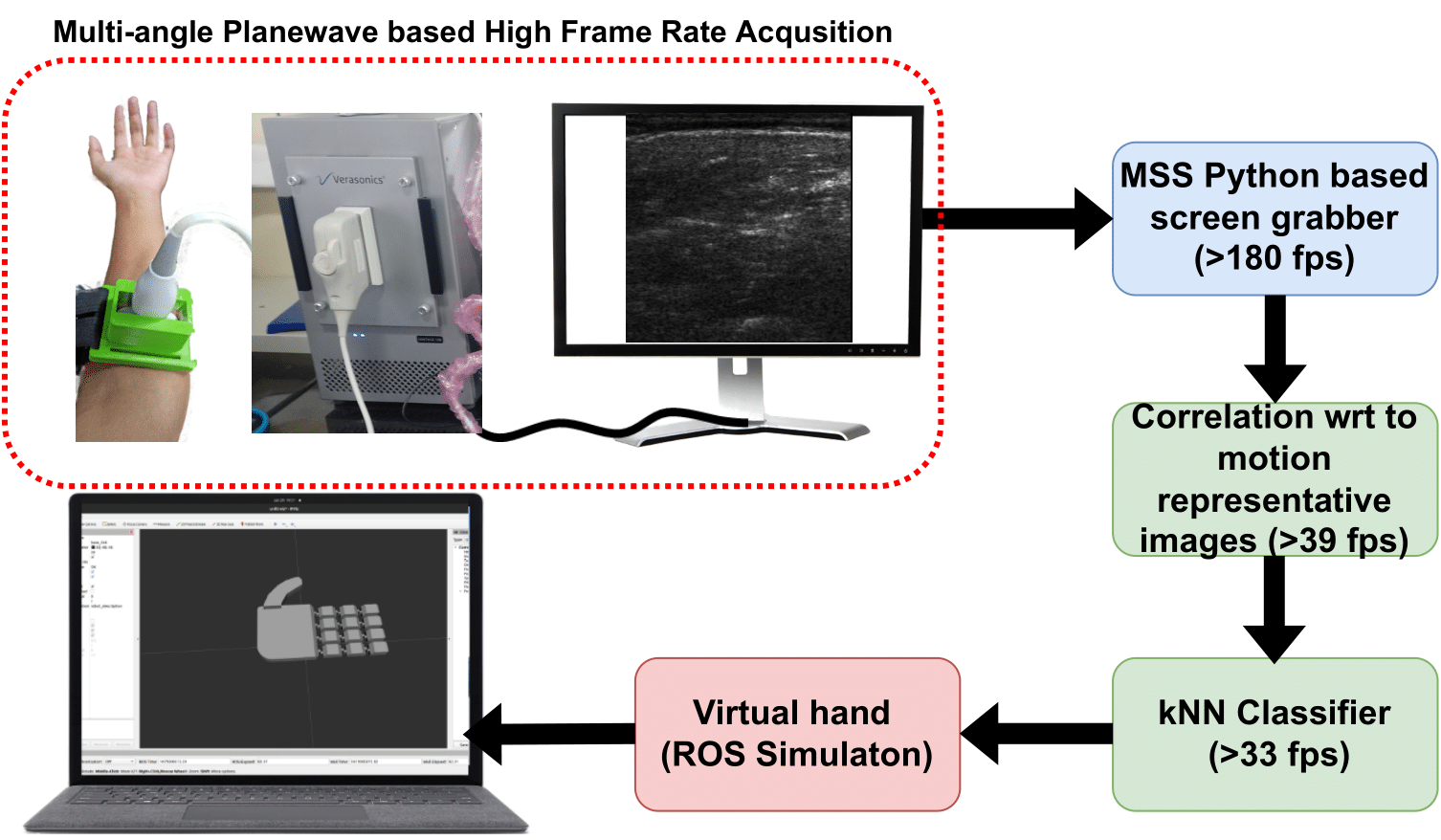}
    \caption{Proposed methodology. The data acquired using a 128 channel research ultrasound platform are screen grabbed and processed for correlation based detection of reference images. The identified reference images are passed through a kNN classifier to detect the four types of hand positions as in Fig. \ref{fig:handgestures}}
    \label{fig:proposedmethodology}
\end{figure*}

\section{METHODOLOGY}
Sonomyography has been an area of significant research in the recent past. Towards this, in this work, a novel prototype system for real-time arm control feedback with the help of a high frame rate ultrasound image acquisition and the ROS visualization (Rviz) is presented \cite{ros}. The proposed framework is as shown in Fig. \ref{fig:proposedmethodology}. A high frame rate ultrasound acquisition ($>500$ fps) using multi-angle planewave excitation employing a 128 channel Verasonics research ultrasound platform was employed for the proposed approach. The beamformed images using the Verasonics software is screen captured using the Python MSS screen grabber tool \cite{pythonmss} with a frame rate of 180 fps. After obtaining the live ultrasound image, the correlation between all representative images and the live image was calculated pixel by pixel as in \cite{dhawan2019proprioceptive} (\ref{eq:pixelcorr}).  
\begin{equation}
    r=\frac{\sum(x_i-\bar{x}).(y_j-\bar{y})}{\sqrt{\sum{(x_i-\bar{x})}^{2}\cdot\sum{(y_j-\bar{y})}{^2}}}
    \label{eq:pixelcorr}
\end{equation}
where $x_i$ belongs to the $i^{th}$ pixel of image 1, and $y_j$ belongs to $j_{th}$ pixel of image 2. 
$\bar{x}$ and $\bar{y}$ refer to the mean of pixel values. The one with the highest correlation was designated as the motion attempted by the user. This methodology was employed for both training and validation of the machine learning (ML) model employed (which is the k-Nearest Neighbourhood (kNN) classifier), where users were instructed to execute a predefined motion during training and a random motion during validation. All images were converted to grayscale before doing any operation. The classifier output will form the input to a 14 degree of freedom (DOF) model of a human arm and visualized using the Rviz tool \cite{rviz}.

\subsection{Data Acquisition and Processing}
\begin{figure}[ht]
    \centering
    \includegraphics[width=\linewidth]{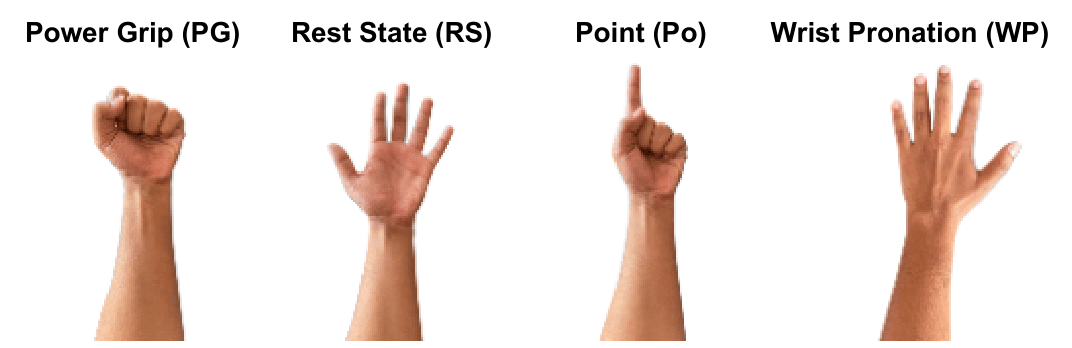}
    \caption{The four classes of hand gestures in this work adapted from \cite{dhawan2019proprioceptive}.}
    \label{fig:handgestures}
\end{figure}

\begin{figure*}[ht]
    \centering
    \includegraphics[width=0.9\linewidth]{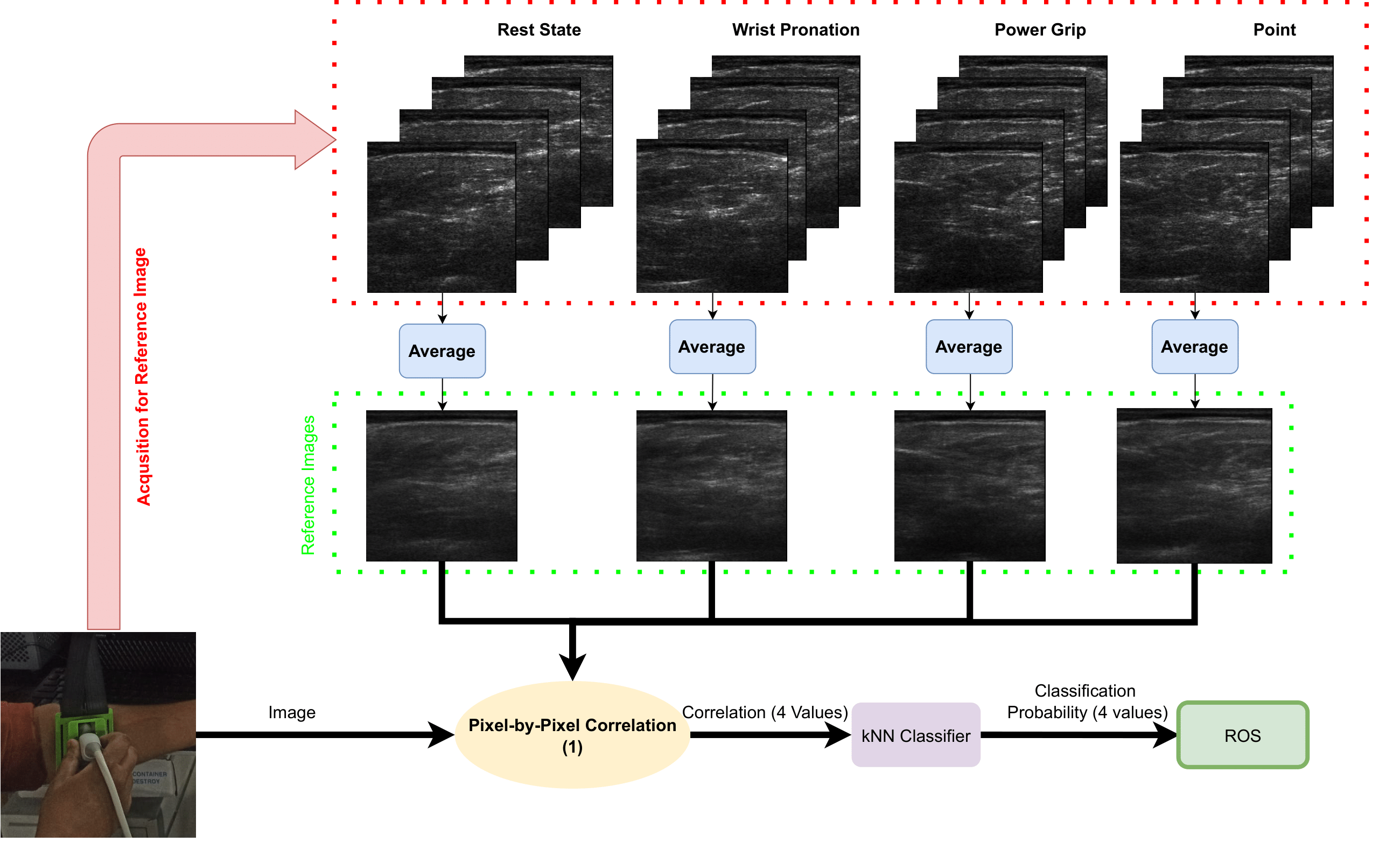}
    \caption{The processing pipeline which consists of the initial acquisition for creating the reference images for each of the four gestures using averaging of images, pixel-by-pixel correlation for each of the acquired frame with the reference images to create correlation feature which will be input to the kNN classifier.}
    \label{fig:processing_pipeline}
\end{figure*}

Following probe placement, the participants went through an initial training phase in which they were asked to hold a given hand gesture for a set period of time ($\sim10s$) before changing the motion to another. As indicated in Fig \ref{fig:handgestures}, we have concentrated on four hand gestures/motions for this work: power grip, wrist pronation, point, and rest state out of the various options presented in \cite{dhawan2019proprioceptive}. The processing pipeline is detailed further in Fig. \ref{fig:processing_pipeline}. The best scans based on visual assessment were then combined to create a single representative image by averaging and labelling each hand gesture as shown in the first row of Fig. \ref{fig:processing_pipeline}. This image was considered to be the reference image for later use for the pixel-by-pixel correlation. Following the processing pipeline, participants were instructed to perform a specific action from the four specified motions while receiving a live ultrasound scan from the hand. 

\subsection{ROS Simulation Details} 
\begin{figure}[ht]
    \centering
    \includegraphics[width=0.5\columnwidth]{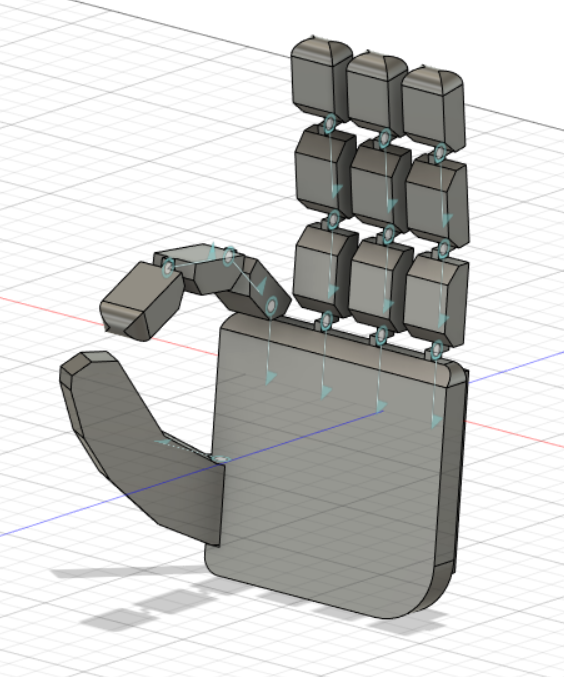}
    \caption{CAD Model of the robotic hand}
    \label{fig:cadmodel}
\end{figure}

In this work, a 14DOF CAD model of a robotic arm that mimicks a human hand was designed as shown in Fig. \ref{fig:cadmodel}. The CAD model was used to generate Unified Robotics Description Format (URDF). The URDF describes to ROS the physical features of a robot. ROS then uses these files to inform the computer about the robot's appearance. This model was loaded using ROS visualisation provided by Rviz \cite{rviz}. Each joint was assigned a radian angle value, and changing the angles resulted in the desired motion. The angles corresponding to power grip, wrist pronation, rest state, and point were hard coded. The virtual hand is designed in such way that it will start moving as soon as the motion was predicted by the kNN classifier. A $0.6s$ time latency was noticed. The complete system setup is as shown in Fig. \ref{fig:holder_hand}(c) and video demonstration of the same is available in \cite{viddemo} for reference.

\subsection{Experimental Details}
\begin{figure}[ht]
    \centering
    \includegraphics[width=0.9\linewidth]{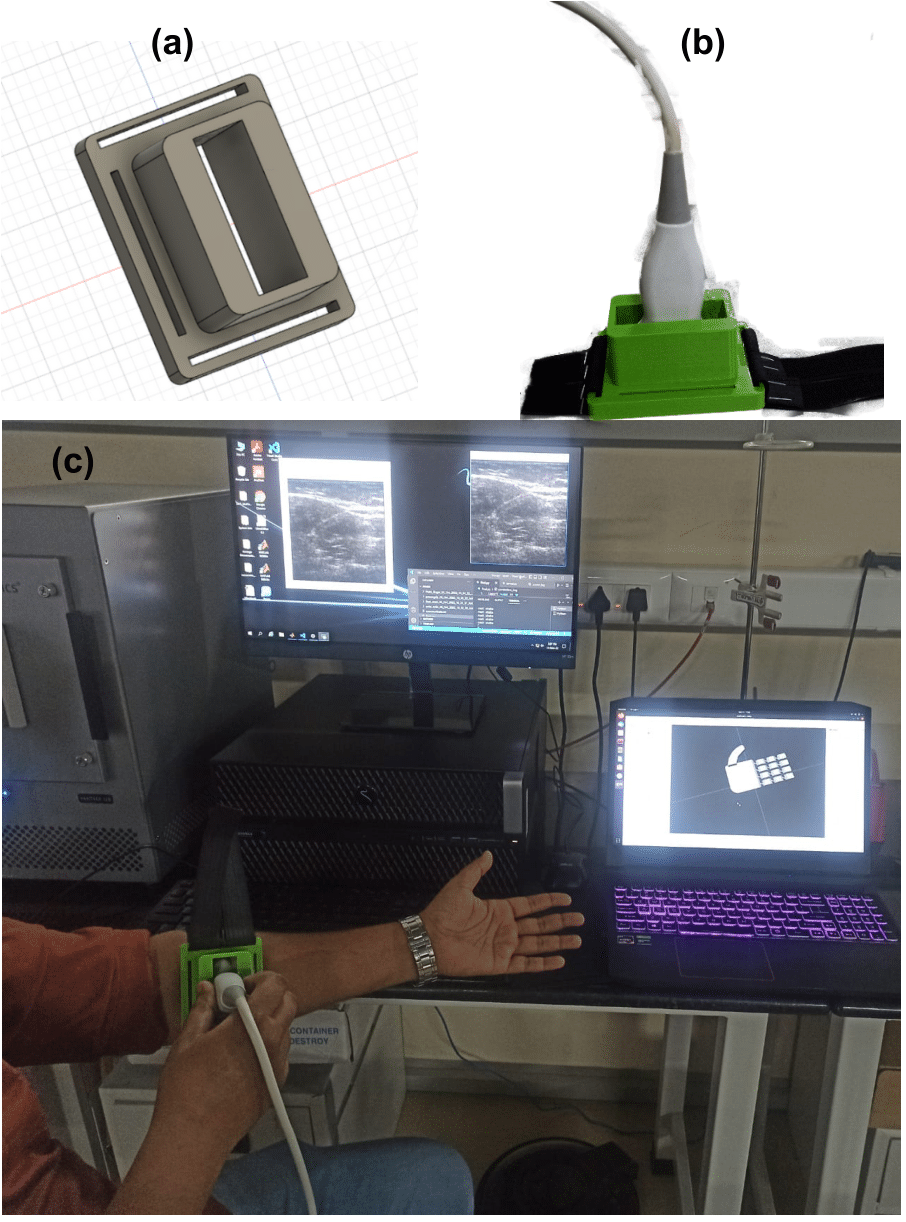}
    \caption{Custom probe holder design. (a) Fusion 360 model adapted to Verasonics L11-5v probe, (b) 3D printed design with tightening belt and connected probe, (c) the complete data acquisition prototype.}
    \label{fig:holder_hand}
\end{figure}

In this work, the data was acquired from three healthy participants. The participant information are briefed in Table \ref{tab:subjectinfo}. All of the experiments described in this work were approved by the ethics committee of Indian Institute of Technology, Palakkad, and they were carried out in accordance with applicable standards and regulations. Before participating in the study, all individuals provided written, informed consent.
\begin{table}
\centering
\caption{Subject Information}
\begin{tabular}{c|c|c|c|c}
\hline
\textbf{Subject ID} & \textbf{Age} & \textbf{Sex}  & \textbf{Weight (Kg)} & \textbf{Dominant Arm} \\ \hline \hline
\textbf{ID1}        & 21  & Male & 84          & Right        \\ \hline
\textbf{ID2}        & 19  & Male & 51          & Right        \\ \hline
\textbf{ID3}        & 36  & Male & 94          & Right        \\ \hline
\end{tabular}
\label{tab:subjectinfo}
\end{table}
The participants were asked to sit upright with their arms propped up for the duration of the trial. The ultrasound probe L11-5V, which was connected to the Verasonics system, was then fastened to the forearm using a custom designed holder. The transducer was manually positioned in the volar aspect around 4-5 cm from the elbow joint as seen in Fig \ref{fig:holder_hand}. This position was chosen so that the deep and superficial flexor muscle compartments could be scanned \cite{dhawan2019proprioceptive,engdahl2022first}.

The python MSS frame grabber \cite{pythonmss} was employed to capture the screen. The captured image was downsized and cropped to contain only the relevant ultrasound image ($480\times480$ pixels) as a frame rate greater than 35 fps. Using the cross-correlation algorithm followed by a kNN classifier, the collected image frames were then analysed for predictions. The Verasonics host system was also linked to a laptop to simulate virtual robotic hands through the TCP/IP Ports. The ROS (Robot Operating System) \cite{ros} was used to perform the simulation, which is the de-facto standard in robotics for creating robot applications.

\subsection{Experimental task on motion discriminability}
The goal of this experiment was to see how successfully our system differentiated between different motions made by participants. Throughout the experiment, participants were shown ultrasound images of their limbs as well as the simulation of the virtual hand in ROS, as detailed in the subsequent section. This study included all of the subjects indicated in Table \ref{tab:subjectinfo}. Every participant did power grip, wrist pronation, rest, and point repetitions. Initially, participants were obliged to incorporate a rest state between transitions from one gesture to another. Following that, they were told to choose any of the gestures at random, without including the rest state through the transition.

\section{Results and Discussions}
\begin{figure}[ht]
    \centering
    \includegraphics[width=0.75\linewidth]{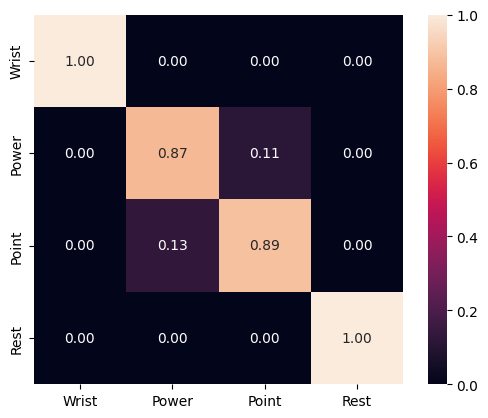}
    \caption{Post-training motion discriminability of subjects as indicated by aggregate confusion matrices}
    \label{fig:confusion}
\end{figure}

The proposed prototype system was tested on healthy subjects. Given the less number of tested subjects (two of the three people) a 100\% cross validation accuracy was achieved across all four activities (during the training). The aggregate confusion matrix for all people is depicted in Fig. \ref{fig:confusion}. It demonstrates that two of four actions were successfully anticipated with 100\% accuracy, however the point was wrongly projected as Power Grip in other circumstances. We also investigated the effect on prediction accuracy of treating rest as a distinct motion class rather than as the absence of motion (low similarity to motion frames). When rest was removed from cross-validation, the motion discriminability of all people increased, suggesting that the classifier could accurately predict motions when presented with representative image frames repeated across numerous trials. The future work involves comparison of the proposed approach with sEMG and also improving on the classification accuracy with more number of subjects. The proposed approach can also be extended to the complete hand gestures as illustrated in \cite{dhawan2019proprioceptive}.
\section{Conclusion}
We tested an ultrasound-based approach for detecting finger movement in response to forearm muscle deformations in this study. The amplitude resolution of conventional myoelectric control based on dry, surface electrodes is limited. Therefore In this study, we employed sonomyography to measure individuals' capacities to modulate end-effectors in real time. In our current study, the 128 channel Verasonics research ultrasound platform was employed to collect the data from the participants' forearms. The scanned frames are provided to a trained kNN classifier to predict the hand gesture and provided as an input to an ROS visualization tool. A 14DOF CAD model is employed in the RViz tool to visualize the desired hand gesture. To employ this control strategy in practise, the size of the ultrasound system and other electronics must be reduced. Another challenge with sonomyography is the movement of the transducer. It has been discovered that increasing the number of training samples at different arm placements can lessen the effect on classification accuracy. Minor variations in transducer location and orientation induced by residual movement can be reduced in real-time utilising proper filtering and machine learning techniques. This work shows that an ultrasound-based, noninvasive muscle activity monitoring technology can be used to control prosthetic arms in real time by individuals with trans radial amputation. The proposed prototype system allows for intuitive control of future prosthetic devices, potentially reducing device rejection.








\bibliographystyle{IEEEtran}

\begin{thebibliography}{99}
\bibitem{biddiss2007upper}
E.~A. Biddiss and T.~T. Chau, ``Upper limb prosthesis use and abandonment: a
  survey of the last 25 years,'' \emph{Prosthetics and orthotics
  international}, vol.~31, no.~3, pp. 236--257, 2007.

\bibitem{clancy2002sampling}
E.~A. Clancy, E.~L. Morin, and R.~Merletti, ``Sampling, noise-reduction and
  amplitude estimation issues in surface electromyography,'' \emph{Journal of
  electromyography and kinesiology}, vol.~12, no.~1, pp. 1--16, 2002.

\bibitem{ortenzi2015ultrasound}
V.~Ortenzi, S.~Tarantino, C.~Castellini, and C.~Cipriani, ``Ultrasound imaging
  for hand prosthesis control: a comparative study of features and
  classification methods,'' in \emph{2015 IEEE International Conference on
  Rehabilitation Robotics (ICORR)}.\hskip 1em plus 0.5em minus 0.4em\relax
  IEEE, 2015, pp. 1--6.

\bibitem{dhawan2019proprioceptive}
A.~S. Dhawan, B.~Mukherjee, S.~Patwardhan, N.~Akhlaghi, G.~Diao, G.~Levay,
  R.~Holley, W.~M. Joiner, M.~Harris-Love, and S.~Sikdar, ``Proprioceptive
  sonomyographic control: A novel method for intuitive and proportional control
  of multiple degrees-of-freedom for individuals with upper extremity limb
  loss,'' \emph{Scientific reports}, vol.~9, no.~1, p. 9499, 2019.

\bibitem{engdahl2022first}
S.~M. Engdahl, S.~A. Acu{\~n}a, E.~L. King, A.~Bashatah, and S.~Sikdar, ``First
  demonstration of functional task performance using a sonomyographic
  prosthesis: a case study,'' \emph{Frontiers in Bioengineering and
  Biotechnology}, vol.~10, 2022.

\bibitem{bercoff2011ultrafast}
J.~Bercoff, ``Ultrafast ultrasound imaging,'' \emph{Ultrasound imaging-Medical
  applications}, pp. 3--24, 2011.

\bibitem{pythonmss}
``{Python MSS Frame Grabber}.'' [Online]. Available:
  \url{https://python-mss.readthedocs.io/}


\bibitem{ros}
``{Robotic Operating System}.'' [Online]. Available: \url{https://www.ros.org/}


\bibitem{rviz}
``{ROS Visualization}.'' [Online]. Available: \url{http://wiki.ros.org/rviz}

\bibitem{viddemo}
``{Video Demo}.'' [Online]. Available: \url{https://drive.google.com/file/d/1Y3Ndd4yfSsTjUxKCgSY4S_W5_1LOeorf/view}

\end{thebibliography}

\end{document}